# KIRA: A Full-Stack Reading Screening System with Dual-Pathway Analysis for Children

Maggie Dong*, Rohit Mekkoth*, Zixuan Wang*, Cheol Jun Cho†, Jiachen Lian†, and Gopala K. Anumanchipalli
*Dept. of Electrical Engineering and Computer Sciences*
*University of California, Berkeley*
Berkeley, CA, USA
{mdong15, rohit_mekkoth, bonniew26, cheoljun, jiachenlian, gopala}@berkeley.edu

***Abstract*—Early-grade reading screening is critical for identifying children at risk of reading difficulty. Existing AI-based screeners are often limited in the scope of their analyses and evaluation metrics, or are proprietary systems that lack transparent model specifications and evaluation. We present KIRA, a dual-pathway screening system that pairs word-level fluency scoring with detailed phoneme-level error detection. We adapt pretrained word and phoneme recognizers to children's speech, reducing word error rate from 13.88% to 8.84% and phoneme error rate from 41.66% to 23.00%. Powered by these child-adapted models, we implement Word Reading (WRE) and Oral Reading Fluency (ORF), two foundational tasks in K–2 screening [1, 2]. The system integrates these recognizers with a web-based proctor-facing interface presenting fluency scores alongside phoneme feedback for potential targeted intervention. KIRA brings modern child-speech recognition into a full-stack, deployable, classroom-ready reading screener.**

***Demo Link***: *https://tinyurl.com/kira-demo*

## I. Introduction

Foundational reading proficiency is a global concern: an estimated two-thirds of ten-year-olds cannot read and understand a simple written story [3]. Because early identification is essential for intervention, governments worldwide have heavily promoted early-grade reading screening [4, 5, 6].

Recent AI-based reading assessment work has begun to make early-grade screening scalable [7, 8]. However, existing speech recognition models are trained primarily on adult speech and transfer poorly to children's speech [9, 10], yielding accuracy below what screening requires. Most prior work has focused on recognition or scoring components [7, 8], rather than integrating them into a teacher-facing system with audio capture and reporting.

Beyond recognition accuracy, a screening system must surface intervention-relevant errors. Word-level scoring, the standard output of most screeners, reports whether a word was read correctly but not how it was misread. A child who reads *"dog"* with a substituted vowel is still marked correct, hiding a pronunciation pattern clinicians would flag. Prior work in disfluent speech modeling shows that accurate clinical assessment requires transcription at both the word and phonetic levels [11, 12, 13]. Meaningful screening therefore requires both fluency scoring and sub-word error analysis.

We present KIRA, Kids' Integrated Reading Analysis, a full-stack screening system integrating dual-pathway speech recognition: word-level fluency scoring alongside parallel phoneme-level analysis with a teacher-facing interface. Both models are fine-tuned on children's speech for improved reliability over baseline adult-oriented speech models. We implement Word Reading (WRE), which assesses isolated-word decoding, and Oral Reading Fluency (ORF), which assesses connected-text reading fluency. These are foundational skills that predict later reading proficiency and are widely used in K–2 screening batteries [1, 2]. Our contributions are:

1) **Word- and phoneme-level scoring in a single pipeline.** KIRA pairs word-level fluency scoring with parallel phoneme-level error attribution, giving teachers both the battery-standard score and sub-word detail for diagnosis and intervention.
2) **Full-stack pipeline for reading screening.** An end-to-end pipeline integrating speech models with a teacher-facing report interface for reading tasks widely adopted in K–2 screening.

## II. Related Work

**Automated Oral Reading Fluency Assessment**. Prior academic systems have focused on the word level, from early tutors [14] to more recent efforts [7, 15]. A separate body of work has focused on phoneme-level analysis: K-Function [8] pairs phoneme transcription with LLM-driven scoring for children's English, and Dudek et al. [16] present a phoneme-level screening pipeline for Polish-speaking children. Commercially, systems have deployed AI-based reading assessment at classroom scale [17, 18], but their recognition architectures, training methods, and benchmark evaluations remain undisclosed, limiting reproducibility and independent evaluation while risking misidentification of at-risk readers.

**Children's ASR**. Adapting modern speech models to children's speech is an active research area, spanning benchmarking on child speech [10], adaptation for classrooms and non-native settings [19], and direct model adaptation via Whisper fine-tuning [9]. KIRA builds on this line of work, fine-tuning modern word- and phoneme-level recognizers on children's speech [20].

*These authors contributed equally to this work.
†Co-project leads.

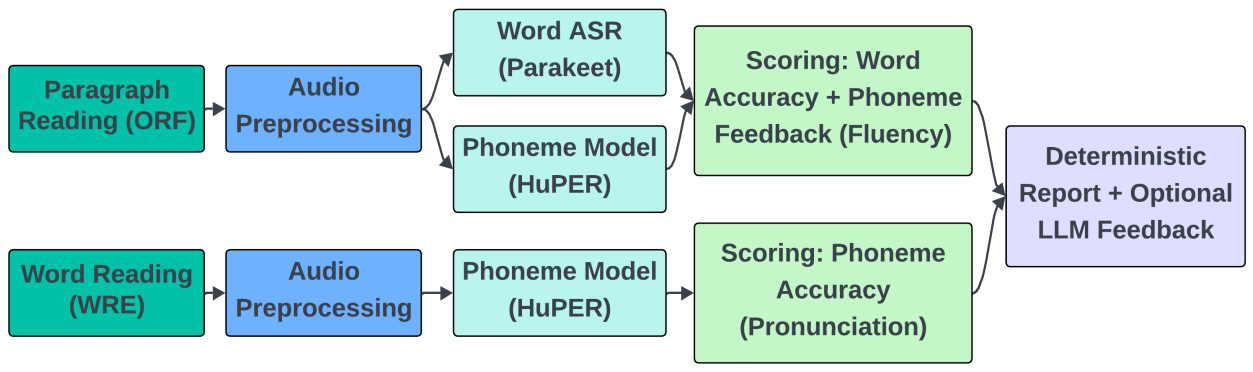

Fig. 1: KIRA pipeline: ORF combines word- and phoneme-level recognition; WRE uses phoneme recognition alone. Both produce a deterministic report with optional LLM feedback.

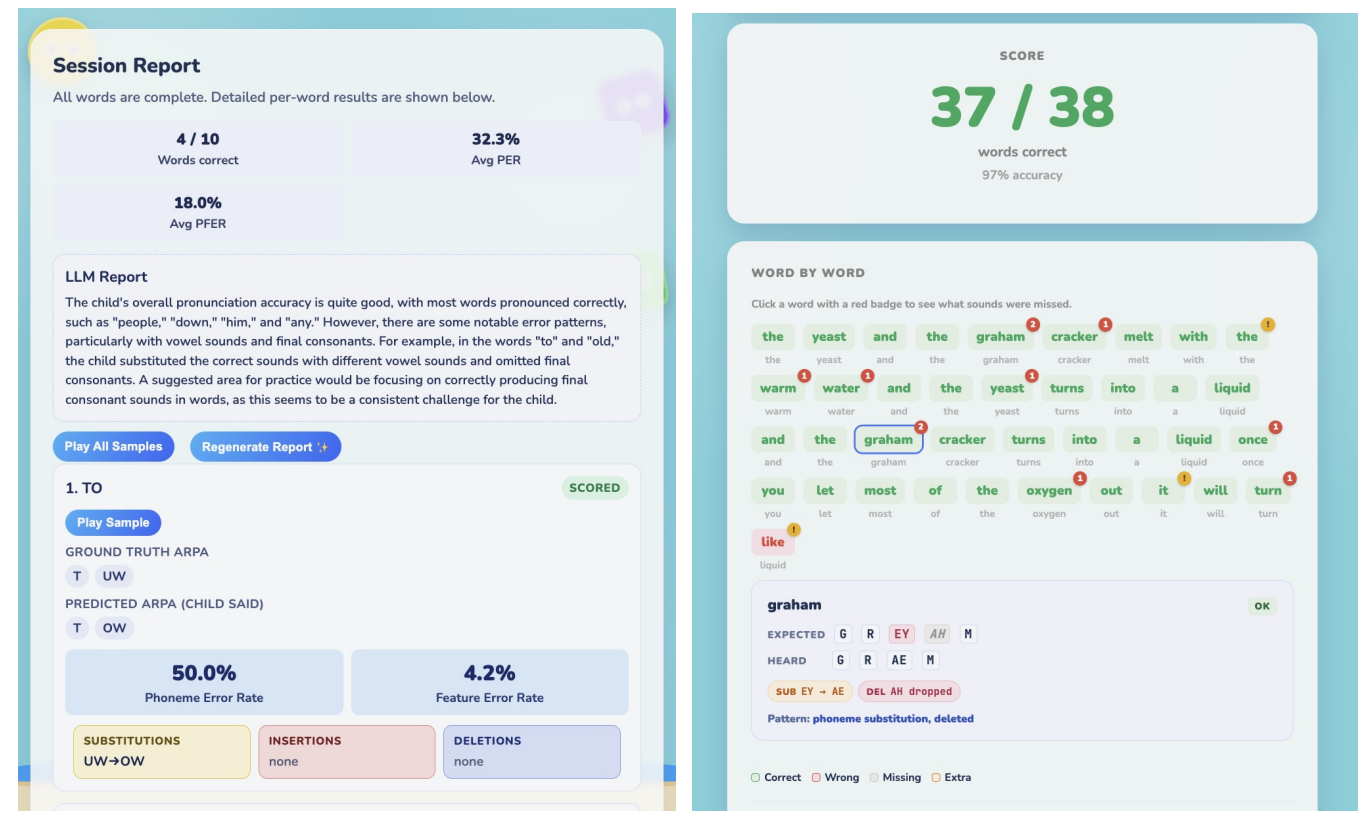

(a) WRE scored output (b) ORF scored output

Fig. 2: KIRA task interfaces: (a) WRE PER and PFER with LLM feedback (b) ORF word-level accuracy and grid. Expanding interfaces shows heard phonemes and labeled errors.

## III. System Description

KIRA is a modular speech-assessment system with three stages. An application layer captures the child's audio and coordinates the pipeline (III-F). Two recognition models (III-A) analyze that audio at complementary granularities: a word-level model produces the transcript used for fluency scoring, and a phoneme-level model produces the sub-word error details. A scoring stage and shared report stage turns their output into the metrics and feedback a user reads (Fig. 1). Before recognition, incoming audio passes through a preprocessing stage. A quality gate rejects unusable recordings, and word-level tasks additionally use Silero VAD [21] to segment the speech regions. KIRA supports two tasks from a battery used in early reading and language screening [2]: ORF (III-C), in which the child reads a connected passage aloud, and WRE (III-B), in which the child reads an isolated word.

### A. Recognition Modules

Both models were fine-tuned on a MyST subset of 20,632 recordings (85/15% train/test) [20], since off-the-shelf models trained on adult speech transfer poorly to children [10]. For word-level transcription, we fine-tuned Parakeet-TDT-1.1B [22], a state-of-the-art ASR model. For phoneme-level transcription, we fine-tuned HuPER [23] on phonemes labeled by applying G2P conversion to the MyST word transcripts, enabling predictions that better reflect spoken form. This property is well-suited to screening, where children's productions often depart from adult norms. HuPER was adapted in two stages: the final CTC layer, then the transformer layers with the low-level feature extractor frozen. In ORF both models run on the same recording; WRE uses only the phoneme model.

### B. Word Reading Pipeline

The WRE pipeline operates on short, single-segment recordings that may contain non-target audio—the proctor speaking, the child sounding out the word, or multiple attempts. Recordings are segmented into voiced regions by Silero VAD [21], which detects speech/silence boundaries. The administration protocol restricts the proctor to a small set of prompts that do not include the target word [2], so speaker separation is handled implicitly by the alignment stage: proctor speech and other non-target audio produce low-scoring alignments.

Each voiced region is aligned to pronunciation variants of the target word using a PanPhon-based feature-weighted edit distance, yielding phonologically-interpretable errors where substitution costs scale with the articulatory-feature distance (e.g., /b/→/p/ costs less than /b/→/k/) [24]. Target items are drawn from a curated list of 170 K–2 sight words from ABC-mouse [25] (e.g., *because*, *always*, *go*). Each word carries 1–2 IPA variants: 165 words have a single canonical pronunciation, and 5 common function words carry a reduced-form alternate (e.g., *a* (/eɪ/ or /ə/), *because* (/bɪkɔz/ or /bɪkʊz/)).

Because a single voiced region may contain both target and non-target audio, we compute two candidate alignments per region: a full-attempt alignment against the complete transcription, and a best-subsequence alignment that finds the cleanest contiguous match. The latter absorbs disfluent sound-out behavior common in K–2 readers ("th-th-the", "buh-buh-bear"). Candidates are classified as full, partial, or irrelevant by reference coverage and token ratio, and the region is scored against the closest variant using the most robust candidate.

### C. Oral Reading Fluency Pipeline

The ORF pipeline runs two parallel analyses on an audio of a child reading a passage: a word-level analysis scores reading accuracy against the reference passage, and a phoneme-level analysis produces per-word diagnostic feedback.

**Word-level:** The word-level model returns a transcript, aligned to the reference passage by longest-common-subsequence matching. Each reference word is labeled correct, missing (unread), or extra (spoken but not in the reference); an adjacent missing/extra pair is collapsed into a single wrong (misread) word. The count of correctly read reference words gives the word-accuracy score (Fig. 2b).

**Phoneme-level:** The phoneme model makes a single forward pass over the whole passage, producing a per-frame distribution over ARPAbet phonemes at 20 ms resolution. The analyzer CTC-decodes this emission into an unconstrained phoneme sequence and aligns it to the reference phonemes, labeling each mismatch as a substitution, deletion, or insertion and attributing it to the reference word it falls under.

Each word is gated by its phoneme coverage, the fraction of its reference phonemes that the recognizer decoded, regardless of correctness. A word left entirely undecoded is marked omitted, a genuine skip; otherwise, coverage below 0.60 flags the word low-confidence and excludes it, so acoustic dropout is not counted as a reading error.

Per-word phoneme results feed into the ORF report, where a flagged word expands to show its expected and heard phonemes with the labeled errors (Fig. 2b).

### D. Scoring

For the phoneme-level evaluation, each error is labeled by type: substitutions (a reference phoneme realized as a different one), deletions (a reference phoneme not produced), and insertions (an extra phoneme), which is surfaced by both pipelines. Word reading also reports a per-word phoneme error rate and phonetic feature error rate (PFER) [24], while paragraph reading reports the number of words read correctly out of the passage total.

### E. Report Generation

Each session assembles these results into a two-level report: a session summary of the per-task metric, and per-word detail that lays out the reference and produced phoneme sequences with the substitutions, deletions, and insertions from §III-D marked in place. Proctors may optionally enable an LLM report layer for privacy-permitting deployments. When on, the scored per-word results are sent to an LLM (GPT-4o-mini) under the clinical system prompt below, returning a narrative summary for both tasks (Fig. 2a).

**Clinical system prompt**: ”You are a speech-language specialist evaluating ORF in K–2 children. Your assessments are read by classroom teachers, not clinicians—use plain language. Focus on phoneme-level error patterns, not general encouragement.”

### F. User Interface

The system is a web application proctors operate directly. They select a task, record or upload the child's speech, and receive the report once analysis finishes. Paragraph reading renders the passage as a word-by-word grid colored by reading outcome, tagging each mispronounced word with the number of phonemes affected. Clicking a tagged word opens its phoneme-level detail card (Fig. 2b). Word reading presents the same per-word detail as a scrollable list, and words with no usable recording are flagged rather than scored. The interface is designed to maximize interpretability for non-specialist proctors through progressive disclosure: the top-level score is shown first, with the sub-word evidence one tap beneath it.

## IV. Evaluation

Because the system's value rests on the accuracy of recognition modules, the performance of those modules are measured on child speech. The test set of MyST is used (3,027 recordings; 16 kHz; mean duration 10–15s; children reading digits, sentences, and science content).

### A. Word ASR

We benchmarked seven pre-trained word-ASR models on MyST, a public child-speech dataset, as a proxy for our target population and fine-tuned the two best-performers by WER (Table I). Parakeet-TDT-1.1B reached 8.84% WER and Parakeet-TDT-0.6B-v2 8.95%; the 0.11% gap from doubling parameters points to a data ceiling rather than a capacity limit. A LoRA-adapted Parakeet-CTC-1.1B trailed at 13.01%, consistent with full fine-tuning being the right strategy at this scale. Our 8.84% is comparable with the strongest published Parakeet result on MyST we found [10], placing the model at a competitive operating point. We therefore deploy Parakeet-TDT-1.1B as the system's word-ASR module.

TABLE I: WER (%) on the MyST test set (3,027 recordings). Bold marks the deployed model.

| Model | Pre-trained | Fine-tuned |
|---|---|---|
| Whisper-Large-v3 | 38.12 | – |
| IBM Granite 3.3-8B | 19.77 | – |
| Wav2Vec2-base-960h | 28.80 | – |
| Canary-Qwen-2.5B | 28.87 | – |
| Parakeet-TDT-0.6B-v2 | 14.91 | 8.95 |
| Parakeet-TDT-1.1B | 13.88 | **8.84** |
| Parakeet-CTC-1.1B | 17.77 | 13.01 (LoRA) |
| Parakeet (Fan et al. [10]) | – | 8.50 |

### B. Phoneme Recognition

We evaluate phoneme recognition (PR) with the same protocol. The base HuPER model [23] scores 41.66% PER on the test set (Table II). This high absolute error reflects the mismatch between its adult-speech pretraining and children's speech. Fine-tuning on children's speech improved performance to 23.00% PER. We therefore deploy the fine-tuned HuPER as the system's PR module.

TABLE II: PER (%) on the MyST test set.

| Model | PER (%) |
|---|---|
| HuPER (base, MyST) | 41.66 |
| HuPER (fine-tuned on MyST) | **23.00** |

## V. Conclusion and Limitations

We presented KIRA, a full-stack pipeline for K–2 reading screening that pairs word-level fluency scoring with parallel phoneme-level error detection, integrated into a web-based proctor-facing interface. KIRA has been validated at the component level; future work will benchmark the phoneme tier against expert labels, compare system- and clinician-written reports, and evaluate diagnostic utility through classroom deployment.

## VI. Acknowledgements

This research is supported by the following grants to PI Anumanchipalli: Robert E. and Beverly A. Brooks Endowed Chair, NIH/NIDCD, and the Schwab Innovation Fund.

## Generative AI Disclosure

The authors used Claude and ChatGPT for manuscript preparation, primarily for grammar and language improvement and condensing all sections to fit the page limit. All content was reviewed and modified by the authors where they deemed necessary, and they take full responsibility for the writeup.